\documentclass{article}
\usepackage{arxiv}
\usepackage[utf8]{inputenc} 
\usepackage[T1]{fontenc}    
\usepackage[hidelinks]{hyperref} 
\usepackage{url}            
\usepackage{booktabs}       
\usepackage{amsfonts}       
\usepackage{nicefrac}       
\usepackage{microtype}      
\usepackage{lipsum}
\usepackage{graphicx}
\usepackage{float}
\usepackage{amsmath}
\usepackage[version=3]{mhchem}
\usepackage{soul}
\usepackage{color}
\usepackage{authblk}
\usepackage{threeparttable}
\usepackage{tabularx}
\graphicspath{ {./Figures/} }
\usepackage{xcolor}
\usepackage{xr}
\newcommand{\floridachem}{Department of Chemistry, University of Florida, Gainesville, FL 32611, USA}
\newcommand{\floridaqtp}{Quantum Theory Project, University of Florida, Gainesville, FL 32611, USA}
\newcommand{\floridamse}{Department of Materials Science \& Engineering, University of Florida, Gainesville, FL 32611, USA}

\newcommand{\nyuphysics}{Center for Soft Matter Research, Department of Physics, New York University, New York 10003, USA}
\newcommand{\nyusimons}{Simons Center for Computational Physical Chemistry, Department of Chemistry, New York University, New York 10003, USA}
\newcommand{\nyucourant}{Courant Institute of Mathematical Sciences, New York University, New York 10003, USA}
\newcommand{\nyucns}{Center for Neural Science, New York University, New York 10003, USA}
\newcommand{\minnesotacems}{Department of Chemical Engineering and Material Science, University of Minnesota - Twin Cities, Minneapolis, MN 55455, USA}
\newcommand{\minnesotaaem}{Department of Aerospace Engineering and Mechanics, University of Minnesota - Twin Cities, Minneapolis, MN 55455, USA}

\begin{document}
\title{MolCrystalFlow: Molecular Crystal Structure Prediction via Flow Matching}

\author[1,2$*$]{Cheng Zeng}
\author[3]{Harry W. Sullivan}
\author[5]{Thomas Egg}
\author[5]{Maya M. Martirossyan}
\author[5,6]{Philipp Höllmer}
\author[1,2]{Jirui Jin}
\author[1,2]{Adrian Roitberg}
\author[2,4]{Richard G. Hennig}
\author[5,6,7,8]{Stefano Martiniani}
\author[9]{Ellad B. Tadmor}
\author[1,2$*$]{Mingjie Liu}
\affil[1]{\floridachem}
\affil[2]{\floridaqtp}
\affil[3]{\minnesotacems}
\affil[4]{\floridamse}
\affil[5]{\nyuphysics}
\affil[6]{\nyusimons}
\affil[7]{\nyucourant}
\affil[8]{\nyucns}
\affil[9]{\minnesotaaem}
\def\thefootnote{*}\footnotetext{c.zeng@ufl.edu, mingjieliu@ufl.edu}

\maketitle
\begin{abstract}
Molecular crystal structure prediction represents a grand challenge in computational chemistry due to large sizes of constituent molecules and complex intra- and intermolecular interactions. While generative modeling has revolutionized structure discovery for molecules, inorganic solids, and metal–organic frameworks, extending such approaches to fully periodic molecular crystals is still elusive.
Here, we present MolCrystalFlow, a flow-based generative model for molecular crystal structure prediction. The framework disentangles intramolecular complexity from intermolecular packing by embedding molecules as rigid bodies and jointly learning the lattice matrix, molecular orientations, and centroid positions. Centroids and orientations are represented on their native Riemannian manifolds, allowing geodesic flow construction and graph neural network operations that respects geometric symmetries. We benchmark our model against a state-of-the-art generative model (MOFFlow) for large-size periodic crystals and a rule-based structure generation method (Genarris) on two open-source molecular crystal datasets. MolCrystalFlow outperforms MOFFlow while achieving competitive performance against Genarris. We also demonstrate an integration of MolCrystalFlow model with universal machine learning potential to accelerate molecular crystal structure prediction, paving the way for data-driven generative discovery of molecular crystals.
\end{abstract}

\section*{Introduction\label{sec:Introduction}}
Molecular crystal structure prediction (CSP) represents a fundamental challenge in computational chemistry.
Unlike inorganic crystals, molecular crystals are governed by an interplay of intramolecular and intermolecular interactions along with periodic lattice constraints, leading to complex energy landscapes with many competing low-energy minima.
This complexity is also known as polymorphism, where a single molecule can adopt multiple crystal packings that can be thermodynamically accessible.
Polymorphism has major implications for drug efficacy, stability, and manufacturability~\cite{buddhadev_pharmaceutical_2021, von_raumer_solid_2018}. A notable example is Ritonavir, an antiretroviral drug initially released in a crystal form known as Form A. Years after commercialization, a previously unknown Form B appeared during routine laboratory analysis~\cite{bauer2001}. Although chemically identical, the two forms differ subtly in molecular packing, leading to drastically different solubility. Form B dissolves poorly in common solvents such as water and methanol, causing drug precipitation and a sharp drop in bioavailability. Its unexpected emergence forced product withdrawals and substantial reformulation costs, demonstrating how failing to anticipate a stable polymorph can have serious clinical and economic consequences.

Predicting stable polymorphs is essential, as applications in optoelectronics~\cite{wei2024_optoelectronic}, pharmaceuticals~\cite{braga2022}, organic semiconductors~\cite{takimiya2024}, and energy storage materials~\cite{muhammad2025} are highly sensitive to slight variations in molecular packing, where small structural differences can result in significant changes in stability and performance. 
Computational approaches have thus become indispensable for polymorph discovery, as experimental screening is slow and resource-intensive.
However, predicting all possible crystal structures for a molecule remains a long-standing challenge in computational chemistry and materials science~\cite{oberan_frontiers_2023}. 
Current CSP workflows rely on stochastic or evolutionary searches to generate candidate structures, followed by large-scale lattice-energy evaluation~\cite{hunnisett2024}. These methods often demand millions of CPU hours for a single compound and are typically system-specific, limiting generalizability across chemical families~\cite{nayal2025}. Machine-learning-based models such as AIMNet2~\cite{nayal2025} and FastCSP~\cite{gharakhanyan2025_FastCSP} accelerate energy prediction but primarily help with ranking; the underlying search over possible packings remains exhaustive and computationally intensive.
Moreover, this prevailing “generate-and-rank” paradigm---relying on brute-force enumeration through stochastic or evolutionary sampling---produces large pools of candidate structures that must subsequently be deduplicated, clustered, and filtered~\cite{yang2025, curtis_gator_2018} before stability rankings are computed. These ranking, typically obtained by DFT-based screening~\cite{gharakhanyan2025_FastCSP, oberan_frontiers_2023} or machine learning models, continue to face a fundamental bottleneck arising from the sheer size of the search space. More efficient and generalizable strategies are needed to guide exploration of the polymorph landscape without exhaustive sampling.

Generative modeling is a promising alternative approach to the brute-force approach using random structure search. It learns a mapping between an easy-to-sample base distribution and a target distribution~\cite{lipman2023, albergo2023, liu2022, song2021, du2024machine, zeng_d_vae_2024}, often using neural networks that respect geometric symmetry constraints. 
A generative model trained on known stable polymorphs might directly propose plausible crystal packings, overcoming kinetic barriers that are normally inaccessible in conventional atomistic simulations. 
Recent advances in generative AI have demonstrated potential in accelerating the prediction and discovery of small molecules~\cite{hoogeboom2022, zeng_propmolflow_2026, jin_molguidance_2025}, protein sequences~\cite{campbell2024}, and inorganic crystals~\cite{hoellmer2025, zeni2025} using diffusion models and flow-based generative processes. 
However, extending these methods to fully atomistic, periodic molecular crystals is non-trivial.
Single molecules lack the periodic constraints intrinsic to molecular crystals, whereas all-atom inorganic crystal models suffer from scaling up.
For instance, given the compositions and numbers of atoms in the crystal, our recent model OMatG accurately matches $\sim$70\% of the ground-truth crystal structures with up to 20 atoms but drops to only 27.4\% at sizes near 50 atoms~\cite{hoellmer2025}. 
This decline reflects the difficulty of learning long-range, many-body interactions, enforcing periodic symmetry, and navigating the vast configuration spaces inherent to larger crystalline systems. 

To address complexity in larger systems, recent cross-scale approaches, such as MOFFlow for metal-organic frameworks (MOFs)~\cite{kim2025} and AssembleFlow for molecular clusters~\cite{guo2024}, represent materials as assemblies of rigid-body building blocks and learn their spatial arrangements. 
These models are promising for framework materials and supramolecular architectures, yet they are not directly applicable to molecular crystals because molecular clusters modeled in AssembleFlow do not naturally exhibit periodic lattices, and MOFFlow does not enforce periodic translational invariance.  
Generative models for molecular crystal structure prediction are just emerging---Jin et al.~\cite{jin2025oxtal} convert the periodic molecular crystal structure prediction challenge to learning the packing for non-periodic molecular clusters with explicit long-range inter-molecular interactions.
All-atom coordinates are then generated using non-equivariant Alphafold3-style transformer backbones. 
However, energies cannot be directly evaluated on the generated clusters without post hoc lattice inference, limiting the utility for downstream applications. 
Generative modeling for molecular crystals remains an open frontier. 
In addition, challenges remain for large structure sizes and complex energy landscapes, and 
no existing generative model can produce stable and diverse molecular crystal polymorphs with periodic lattice constraints. 
This work aims to close this gap by developing a generative framework designed for molecular crystal structure prediction. 
Such an approach will enable more efficient exploration of polymorph landscapes and accelerate the discovery of molecular crystals.

We introduce MolCrystalFlow, a periodic E(3)-invariant flow-based generative model that predicts the crystal packing with explicit lattice periodicity.
MolCrystalFlow disentangles the intramolecular and intermolecular complexity by conditioning the generative process on given molecular conformers that are treated under rigid-body approximation. 
It then generates all modalities of molecular crystals needed to recover the full structures.
Flow matching of molecular crystals operates on the corresponding Riemannian manifolds.
We benchmark MolCrystalFlow on two publicly available datasets, including a small-dataset of 11.5K structures derived from CSD experimental data~\cite{thurlemann2023} and a subset of the largest open-source OMC25 dataset~\cite{gharakhanyan2025_OMC25}. 
We demonstrate superior performance of MolCrystalFlow over MOFFlow~\cite{kim2025}, the state-of-the-art hierarchical flow-based generative models for large-size periodic crystalline materials.
We integrate MolCrystalFlow with universal machine learning interatomic potentials (u-MLIP) and density functional theory (DFT) calculations to perform molecular crystal structure predictions on three open CSP competition targets.
The combined pipeline identifies polymorphs for two target compounds with energies and crystal packing geometries close to experimental structures.
We anticipate that the model and pipeline developed in this work, together with the open data, will enable new research avenues toward end-to-end generative discovery of molecular crystals.

\section*{Results\label{sec:results}}
\subsection*{Overview of MolCrystalFlow}
\begin{figure}
    \centering
    \includegraphics[width=0.95\linewidth]{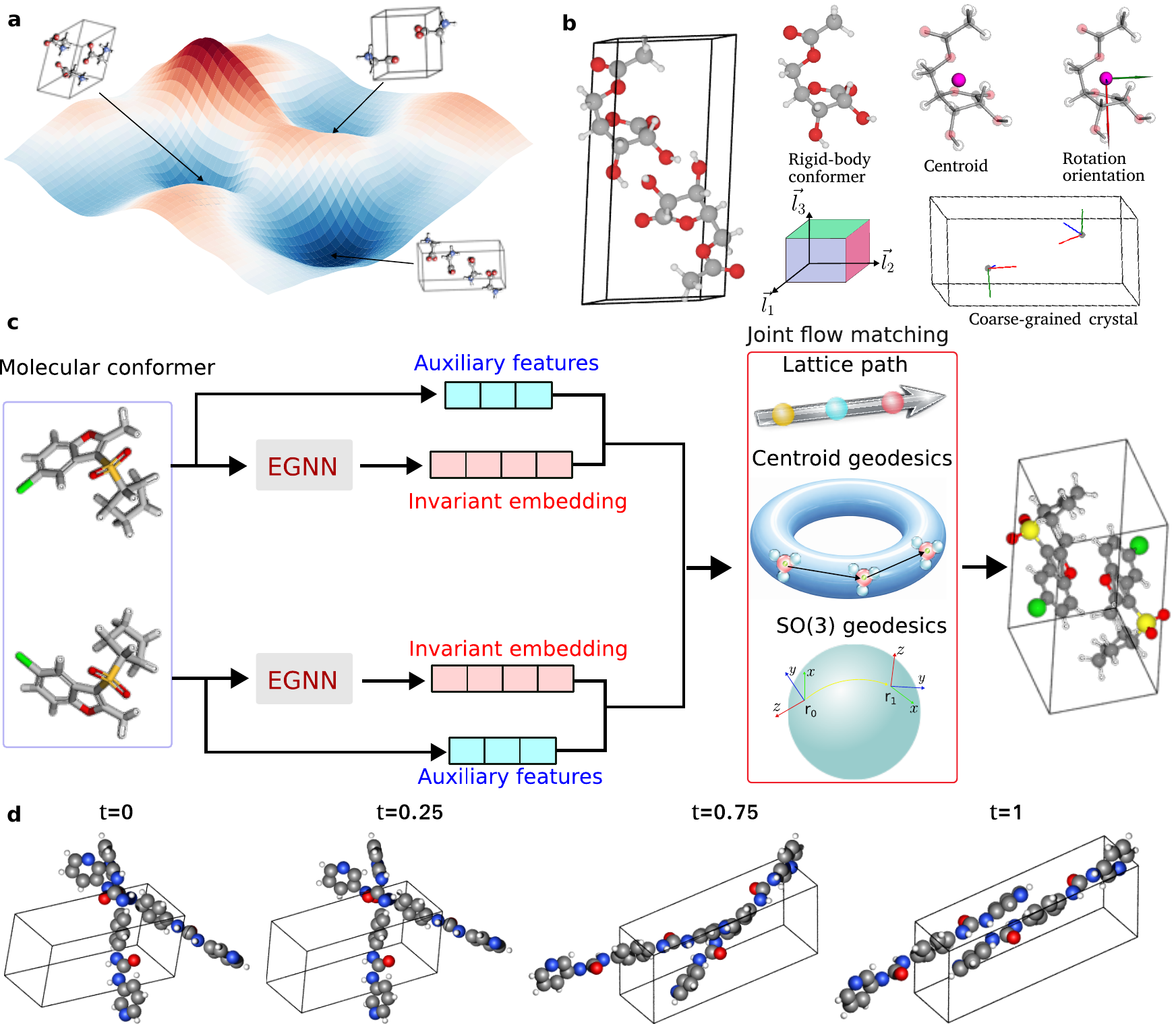}
    \caption{\textbf{Overview of the MolCrystalFlow framework.} a, Schematic illustration of a polymorph energy landscape in molecular crystals. The topology, barrier heights, and relative basin depths are not quantitative and are shown for conceptual purposes only. b, Representation of a molecular crystal structure. A molecular crystal is decomposed to a lattice matrix spanning by three vectors $\overrightarrow{l_1}$, $\overrightarrow{l_2}$ and $\overrightarrow{l_3}$, and a constituent molecular conformer building block which can be characterized by a rigid-body centroid and rotational orientation given by three eigenvectors derived from a principal component analysis over the atomic positions of the building block. Hence a coarse-grained crystal includes all the modalities to be generated. c, MolCrystalFlow works by encoding each building block in the crystal (for example two molecules in the schematic) with an equivariant graph neural network (EGNN) that produces an invariant embedding. The invariant embedding is augmented with auxiliary features to form the input for the joint flow-matching generative process that simultaneously produce the lattice matrix using a linear interpolant for each lattice matrix entry, centroid fractional coordinates on a torus geodesic, and rotational orientation on an $SO(3)$ geodesic. Together, the three learned modalities will reconstruct the full atomic structure of the molecular crystal. d. An example generation trajectory with four time steps, showing the co-evolution of centroid positions, molecular orientations and lattice.}
    \label{fig:overview}
\end{figure}
The overarching goal of molecular crystal structure prediction is to identify low-energy polymorphs on the free-energy landscape (Figure~\ref{fig:overview}a).
Given a molecular graph as input, the task is to predict its crystal packing. 
This problem is typically decomposed into two stages.
First, a stable molecular conformer\footnote{A \textit{molecular conformer} is a specific, 3D spatial arrangement of a molecule's atoms corresponding to a local minimum on the molecule's potential energy surface.} needs to be identified based on the corresponding 2D topological graph, a task for which tools such as OpenEye OMEGA~\cite{hawkins_conformer_2010} and RDKit ETKDGv2~\cite{riniker_better_2015} provide robust solutions.
The subsequent and substantially more challenging step is to predict the crystal packing of these molecular conformers.
Finally, candidate polymorphs are ranked according to their thermodynamic stability, commonly evaluated using DFT-based calculations.

MolCrystalFlow is designed to reduce the search cost and improve the quality of crystal packing by directly sampling physically plausible polymorphs for a given molecular conformer.
To accelerate this search, each molecule within the crystal is represented as a rigid-body conformer, which is a reasonable approximation in many cases~\cite{day2005, jin2025oxtal}. 
Under this assumption, a crystal packing configuration is fully specified by the lattice matrix, along with the centroids and rotational orientations of all molecular rigid bodies (Figure~\ref{fig:overview}b).
The centroid of each molecule is defined as its center of position, while the rotational orientation is parameterized by unit principal axes obtained via principal component analysis (PCA), followed by normalization to ensure consistent and unique local coordinate frames across molecules.

Building on this representation, we adopt a hierarchical modeling framework for molecular crystal structure prediction (Figure~\ref{fig:overview}c).
Similar approaches have been demonstrated in previous generative models for large crystalline MOFs~\cite{kim2025} and molecular clusters~\cite{guo2024}.
In the first stage, each molecular building block is embedded using an equivariant graph neural network (EGNN) based on the E($n$)-equivariant architecture introduced by Hoogeboom et al.~\cite{hoogeboom2022}. The final node features are aggregated to produce an invariant embedding that characterizes the molecular building block. 
Details of this building-block embedder are provided in Figure~\ref{fig:architecture}a. 
Some molecular information---such as size or functional-group complexity---may be attenuated when projected into a neural network latent space. 
As shown in Figure~\ref{fig:umap_bb_embedding}, the latent embedding derived from a trained EGNN model alone exhibits weak correlation with the number of atoms in the building block. 
To enrich the molecular representation, we concatenate 18 auxiliary molecular descriptors to the invariant embedding before passing it to the second stage. 
The full list of these auxiliary features is described in the \hyperlink{sec:representation}{`Hierarchical Representation of Molecular Crystals'} section.

The second stage consists of a joint flow-matching generative process parameterized by an E(3) periodic-invariant neural network, which generates the lattice matrix, molecular centroid positions, and molecular orientations (Figure~\ref{fig:architecture}b). 
Flow matching learns conditional velocity fields that transport samples from a simple base distribution to the target data distribution~\cite{lipman2023, liu2022}. 
By employing a joint flow formulation, distinct interpolants are defined for each modality.
The generative process gradually refines each modality until a final structure is generated.
To construct flows on non-trivial geometries, each modality is defined on its intrinsic Riemannian manifold~\cite{chen2024}. 
Specifically, molecular centroids are represented using fractional coordinates to naturally respect periodic translational invariance, corresponding to a three-dimensional torus. 
Molecular orientations evolve on the $\mathrm{SO}(3)$ manifold. 
We adopt a data-informed base distribution for the lattice matrix and use a simple linear interpolation, which we find to be both stable and effective in practice.
During inference, the generative model simultaneously evolves the lattice parameters, centroid positions, and molecular orientations by integrating the learned velocity fields (Figure~\ref{fig:overview}d). 
By combining the generated lattice matrix, centroid positions and rotational orientations with the internal fixed coordinates of molecular building blocks, MolCrystalFlow reconstructs all-atom molecular crystal structures suitable for direct optimization and energy evaluation.
Details of choices of base distributions and joint flow matching are included in the \hyperlink{sec:base_p}{`Base Distributions and Optimal Transport'} and \hyperlink{sec:joint_flow_matching}{`Joint Flow Matching on Riemannian manifolds'} sections.

\subsection*{Neural Network Architecture}

\begin{figure}
    \centering
    \includegraphics[width=0.95\linewidth]{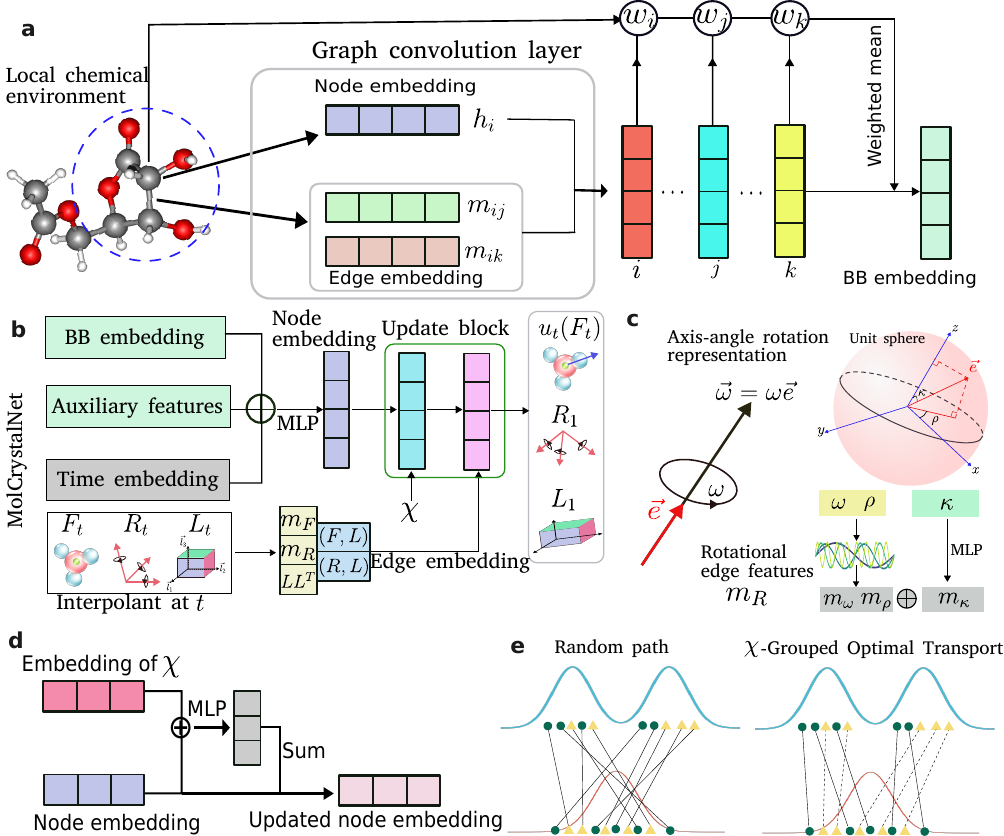}
    \caption{\textbf{Neural network architecture of MolCrystalFlow.} a, Molecular building block (BB) embedding network with EGNN. An atomic node embedding is constructed by the atom interacting with neighbors in the local chemical environment through multiple graph convolution layers. The final building block embedding is a weighted atomic embedding where each weight is learned and comes from the node embedding interacting with invariant atomic distance to the centroid. b, MolCrystalNet parameterizes the joint flow matching for lattice matrix $L$, centroid fractional coordinates ($F$), and rotational orientation ($R$). Given interpolants at time $t$, a concatenation of BB embedding, auxiliary features and time embedding, followed by a multilayer perceptron (MLP), gives rise to the initial node embedding for each molecular building block. In the update block, the initial embedding interacts with a $\chi$ embedding for each building block. Message passing relies on the relative positions and orientations between building blocks, and current lattice matrix at time $t$. The final output is the velocity field for centroid fractional coordinates ($u_t(F_t)$), the denoised rotational orientation ($R_1$) and lattice matrix ($L_1$). Interactions of lattice with fractional coordinates $(F,L)$ and orientations $(R,L)$ are also included in the message passing. c,  For rotation-dedicated message passing ($m_R$), we use the axis-angle representation of relative rotational orientation between two building blocks. We embed the rotational angle $\omega$ and azimuthal angle $\rho$ using Fourier-series like trigonometric functions while using an MLP to create the embedding for the inclination angle $\kappa$. d, Concatenation-summation operation to strengthen the signal of each building block's $\chi$. e, $\chi$-grouped optimal transport to reduce cross-links between paths of different building blocks and facilitate inference.}
    \label{fig:architecture}
\end{figure}

As a rigid body, the representation of each individual molecular building block must be invariant to global translations and rotations. Accordingly, the embedding is designed to capture the intrinsic chemical and geometric structure of the molecule, independent of its absolute position or orientation in space. The global placement of molecules—namely, their centroid positions and rigid-body orientations within the crystal lattice—is learned in a subsequent stage of the model.
To obtain an E(3)-invariant molecular representation, we employ an EGNN (Figure~\ref{fig:architecture}a). Each atom $i$ is initialized with an atom-type embedding and its Cartesian coordinates. Node embeddings $h_i$ are updated through message passing $m_{ij}$ between atoms within a cutoff radius, which depend on interatomic distances. Therefore, the resulting node embeddings are invariant to global translations and rotations. The molecular building-block (BB) embedding is then computed as a learnable weighted mean of the final node embeddings, where the weights $w_i$ are predicted by a shallow MLP using the final node features and each atom’s distance to the molecular centroid.
Details of the BB embedding network are provided in the \hyperlink{sec:representation}{`Hierarchical Representation of Molecular Crystals'} section.

Given the molecular building-block embeddings, the remaining task is to generate the crystal lattice matrix, fractional centroid positions, and rigid-body orientations. To this end, we employ a periodic E(3)-invariant graph neural network adapted from DiffCSP~\cite{jiao2024} that explicitly respects the physical symmetries governing molecular crystal packing, including periodic translational invariance, equivariance of lattice vectors, and equivariance of molecular orientations under global rotations (Figure~\ref{fig:architecture}b).
Geometric symmetries of molecular crystals are discussed in the \hyperlink{sec:geometric_symmetry}{`Geometric Symmetry of Molecular Crystal Structures'} section.
By construction, the network also satisfies permutation invariance with respect to molecules in the unit cell.
The initial node embedding for each building block is formed by concatenating its invariant BB embedding, auxiliary features, and a time embedding~\cite{vaswani_attention_2017, ho_denoising_2020}, followed by projection through a shallow MLP. In the flow-matching framework, the objective is to learn the velocity field $u_\theta(x_t, t)$ associated with an interpolant state $x_t$ at time $t$. 
The interpolants for all structural modalities are described in the \hyperlink{sec:joint_flow_matching}{`Joint Flow Matching on Riemannian manifolds'} section. 
Rather than directly predicting velocity fields for all variables, we adopt a denoising parameterization for selected modalities, as intermediate states are less critical than the final configuration~\cite{dunn2024-1}. Specifically, the loss functions are reparameterized to predict denoised targets for the lattice matrix ($L_1$) and rotational orientations ($R_1$). 
For fractional coordinates, we retain a velocity-field objective ($u_t(F_t)$) to avoid ambiguities arising from periodic wrapping when comparing denoised targets.

Message passing operates on the current states of molecular building blocks. For fractional coordinates, the message $m_F$ is constructed using a Fourier embedding of relative fractional coordinate differences, ensuring periodic translational invariance of centroid positions~\cite{jiao2024}.
To model rigid-body orientations---an essential modality absent in inorganic crystals---we introduce a dedicated rotational message passing term $m_R$. The rotational message depends on relative orientations between building blocks. Each relative rotation matrix is converted to an axis--angle representation $\vec{\omega}$, where the direction vector $\vec{e}$ specifies the rotation axis and the magnitude $\omega$ denotes the rotation angle (Figure~\ref{fig:architecture}c). 
Since $\vec{e}$ has two degrees of freedom, it is parameterized using polar coordinates with an inclination angle $\kappa$ and an azimuthal angle $\rho$. 
Notably, $\omega$ and $\rho$ are periodic with modulo $2\pi$, and are therefore embedded using Fourier features, while $\kappa$ is embedded via an MLP. The resulting embeddings are combined to form the rotational edge feature $m_R$.
In addition, we incorporate interactions of the lattice matrix with fractional coordinate $(F, L)$ and orientation differences $(R, L)$, which represent the product of lattice Gram matrix with the relative fractional coordinates and rotation matrices, following prior work in MatterGen~\cite{zeni2025} and FlowMM~\cite{miller2024}. 
Despite the two-stage encoding of molecular and crystal-level information, the building-block embedder network is trained jointly with  MolCrystalNet.
Details of the periodic E(3)-invariant flow matching and message passing can be found in \hyperref[sec:joint_flow_matching]{`Joint Flow Matching on Riemannian manifolds'} section.

Molecular crystals may adopt packings in which molecules share the same or opposite axis-flip states ($\chi$). 
This feature $\chi$ arises naturally to account for the degeneracy of local reference frames obtained from principal component analyses. 
The formal definition of $\chi$ within the equivariant local frame construction is provided in \hyperref[sec:extract_chi]{`Extracting the axis-flip state $\chi$'} section. 
To account for this additional geometric degree of freedom for each building block, we introduce an element embedding of $\chi$ at each update block, which is fused into the node embedding via a concatenate\_sum operation (Figure~\ref{fig:architecture}c,d).
To mitigate cross-links between flow paths of different building blocks within a crystal, we employ $\chi$-grouped SE(3) optimal transport, which is detailed in the \hyperlink{sec:base_p}{`Base Distributions and Optimal Transport'} section. This design reflects the fact that this axis-flip state is an intrinsic property of each building block and cannot be altered through rotational evolution during the generative process (Figure~\ref{fig:architecture}e).

\subsection*{Accurate and fast generation of densely packed molecular crystals}

\begin{figure}
    \centering
    \includegraphics[width=0.95\linewidth]{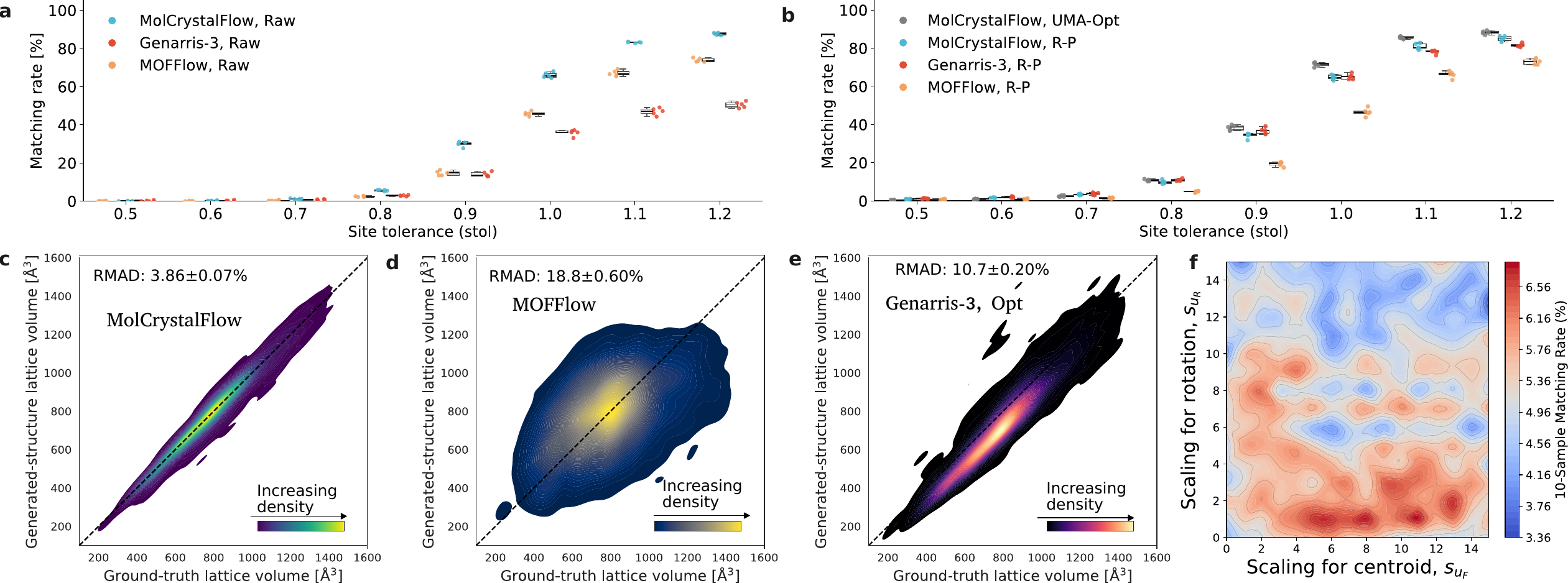}
    \caption{\textbf{Performance of MolCrystalFlow against MOFFlow and Genarris-3 baselines.} a, Comparison of MolCrystalFlow with MOFFlow and Genarris-3 over 10-sample matching rates evaluated for directly generated structures without any optimization under different levels of site tolerances.  b, Comparison of MolCrystalFlow with Genarris-3 and MOFFlow over 10-sample matching rates evaluated for generated structures after a rigid-press optimization (`R-P') under different levels of site tolerances. For MolCrystalFlow, we also show results over samples going through a rigid-body optimization using the UMA-OMC model (`UMA-Opt'). Regarding the box plots, the median is shown as a solid line. The edges of the box correspond to the first and third quartiles, and the whiskers extend to values within 1.5 × interquartile ranges. c--e, Lattice volume deviation of generated structures for MolCrystalFlow (c), MOFFlow (d), and Genarris-3 with rigid-press optimization (e). A kernel density estimation is used to show data distributions for 10 samples. Relative mean absolute deviation (RMAD) is reported as mean and standard deviation across 10 samples.  d, Grid search for the optimal velocity annealing scalings for centroid ($s_{u_F}$) and rotation ($s_{u_R}$) using 10 samples and stol=0.8. For figure a and b, five independent runs for 10 samples are generated the statistics and the results for each run are overlaid on the box plots.}
    \label{fig:benchmark}
\end{figure}

We trained MolCrystalFlow on an open-source molecular crystal dataset curated by Th\"urlemann from CSD~\cite{thurlemann2023}. The dataset comprises 11,488 structures, split to 10,000 training, 738 validation, and 750 test structures.
Further details on data curation and preprocessing are provided in the \hyperlink{sec:datasets}{`Open-Source Molecular Crystal Dataset'} section.
We benchmark MolCrystalFlow against the state-of-the-art generative model MOFFlow designed for large-size periodic crystalline materials under rigid-body approximation such as metal-organic frameworks~\cite{kim2025}, as well as the rule-based Genarris-3 method~\cite{yang_genarris_2025}. 
We do not include fully all-atom generative models designed for periodic inorganic crystals, as these methods struggle to generate physically reasonable structure large crystalline materials~\cite{kim2025}.
We also do not compare to Oxtal~\cite{jin2025oxtal}, which reformulates crystal generation as finite cluster prediction without explicit lattices and is trained on the proprietary full CSD molecular crystal data.
Cluster-based and lattice-explicit models represent complementary formulations of molecular crystal generation. The former emphasizes local packing motifs, while the latter enforces global periodicity. 
In this work, we focus on the periodic formulation with an equivariant graph neural network. This choice ensures that both the model and dataset remain fully open for community development, enables the direct application of standard crystallographic metrics for periodic structures, and allows structure optimization and energy ranking to be performed without requiring any post hoc lattice inference.
We first assess prediction quality using the structure matching rate, a metric widely adopted in the inorganic crystal community~\cite{ong_python_2013}. In addition, we analyze lattice volume deviations between generated and reference structures, which serve as an important indicator of whether densely packed crystals have been identified.

Unlike MolCrystalFlow, MOFFlow operates directly on Cartesian coordinates and generates lattice parameters without enforcing periodic translational invariance.
We retrained MOFFlow on our molecular crystal datasets.
Both MolCrystalFlow and MOFFlow use 50 integration time steps, which is the default setting in MOFFlow~\cite{kim2025} and which we find to be robust for MolCrystalFlow (Figure~\ref{fig:effect_timestep}).
MolCrystalFlow and MOFFlow generate structures in a one-shot manner, whereas Genarris-3 adopts a multi-sample strategy, producing a large candidate pool followed by selection and deduplication to enhance structural diversity. Additionally, MolCrystalFlow and MOFFlow do not impose explicit space-group constraints, while Genarris-3 conditions structure generation on space-group symmetries derived from the molecular point group and the number of molecules per unit cell.
For Genarris-3, we report both the directly generated structures (`Raw') and those after the ``Rigid Press'' optimization (`R-P'), where a hard-sphere potential is applied to refine molecular positions and lattice parameters under space-group constraints.
To enable a fair comparison with these optimized structures, we performed a rigid-press optimization on structures generated by MOFFlow and MolCrystalFlow. 
For MolCrystalFlow, we also used a rigid-body optimization with fixed lattices on MolCrystalFlow generated structures using the UMA-OMC model (`UMA-Opt')~\cite{wood2025, gharakhanyan2025_FastCSP} .
Details of Genarris-3 structure generation and rigid-press optimization are provided in \hyperlink{sec:sec:computational_settings}{`Computational settings'} section.

We first compare structure matching rates across site tolerances ranging from 0.5 to 1.2.
Site tolerance is the maximum allowed positional difference between atoms when deciding whether two crystal structures are considered equivalent~\cite{ong_python_2013}.
For each test compound, 10 samples are generated, and a prediction is deemed successful if at least one sample matches the corresponding ground-truth structure.
This multi-sample evaluation is well motivated, as molecular crystals frequently exhibit polymorphism.
MolCrystalFlow substantially outperforms MOFFlow and Genarris-3 for directly generated structures across all site tolerances (Figure~\ref{fig:benchmark}a).
For rigid-press optimized structures, both Genarris-3 and MolCrystalFlow outperform MOFFlow. 
Genarris-3 achieves higher matching rates under stricter site tolerances, whereas MolCrystalFlow performs better at looser tolerances for both rigid-press and UMA-OMC optimized structures  (Figure~\ref{fig:benchmark}b).
It should be noted that rigid-press improves performance at stricter stol values while decreasing performance at higher stol for MolCrystalFlow.
This trend persists when increasing the number of samples up to 200 (Figure~\ref{fig:effect_n_samples}b,c).
In addition, MolCrystalFlow demonstrates accurate lattice volume prediction.
Across lattice volumes in the range between 200 and 1600 \AA$^3$, the relative mean absolute deviation (RMAD) for all structures of ten samples is 3.86$\pm$0.07\% (Figure~\ref{fig:benchmark}b).
As a comparison, MOFFlow exhibits substantially larger lattice volume deviations of 18.8$\pm$0.6\%, and Genarris-3 shows deviations of 59.0$\pm$0.35\% for directly generated and 10.7$\pm$0.20\% after rigid-press optimization.
Per-structure lattice volume deviation for MolCrystalFlow is detailed in Figure~\ref{fig:per_structure_lattice_dev}.

In Riemannian flow matching, introducing velocity scaling factors on the manifold for fractional coordinates ($s_{u_F}$) and rotational degrees of freedom ($s_{u_R}$) has empirically been shown to improve performance~\cite{hoellmer2025, campbell2024}.
To determine suitable values, we performed a grid search optimizing the 10-sample matching rate at $\text{stol} = 0.8$. 
As shown in Figure~\ref{fig:benchmark}d, performance varies substantially (3.36--6.8\%), with optimal results obtained for $s_{u_F} \in [5, 13]$ and $s_{u_R} \in [1, 3]$.
Unless otherwise stated, Figures~\ref{fig:benchmark}a–c use $s_{u_F}=9$ and $s_{u_R}=3$.
In terms of sampling efficiency, MolCrystalFlow generates structures in an average of 22 ms per structure, approximately twice as fast as Genarris-3 (43 ms per structure).
One should however note that MolCrystalFlow inference runs on a single GPU and 16 CPUs, whereas Genarris-3 operates entirely on 40 CPUs.
While fast in absolute terms, MolCrystalFlow is slower than the non-equivariant transformer-based MOFFlow baseline (6 ms per structure under identical inference settings). 
However, the performance gap between MolCrystalFlow and MOFFlow cannot be offset by increasing the number of samples, as shown in Figure~\ref{fig:effect_n_samples}a.
We further evaluate models trained on a subset of the open OMC25 molecular crystal dataset~\cite{gharakhanyan2025_OMC25}. 
On this larger and more diverse dataset, MolCrystalFlow still outperforms MOFFlow in both structure matching rates (Figure~\ref{fig:omc25_results}) and lattice volume deviations (Figure~\ref{fig:mofflow_omc25}). 
Genarris-3 is not included in the comparison with the OMC25 dataset, as it was used to generate the OMC25 structures.
Details of the selected OMC25 subset are provided in the \hyperlink{sec:datasets}{`Open-Source Molecular Crystal Dataset'} section.

\subsection*{Integrating MolCrystalFlow with u-MLIP for Crystal Structure Prediction}

\begin{figure}
    \centering
    \includegraphics[width=0.9\linewidth]{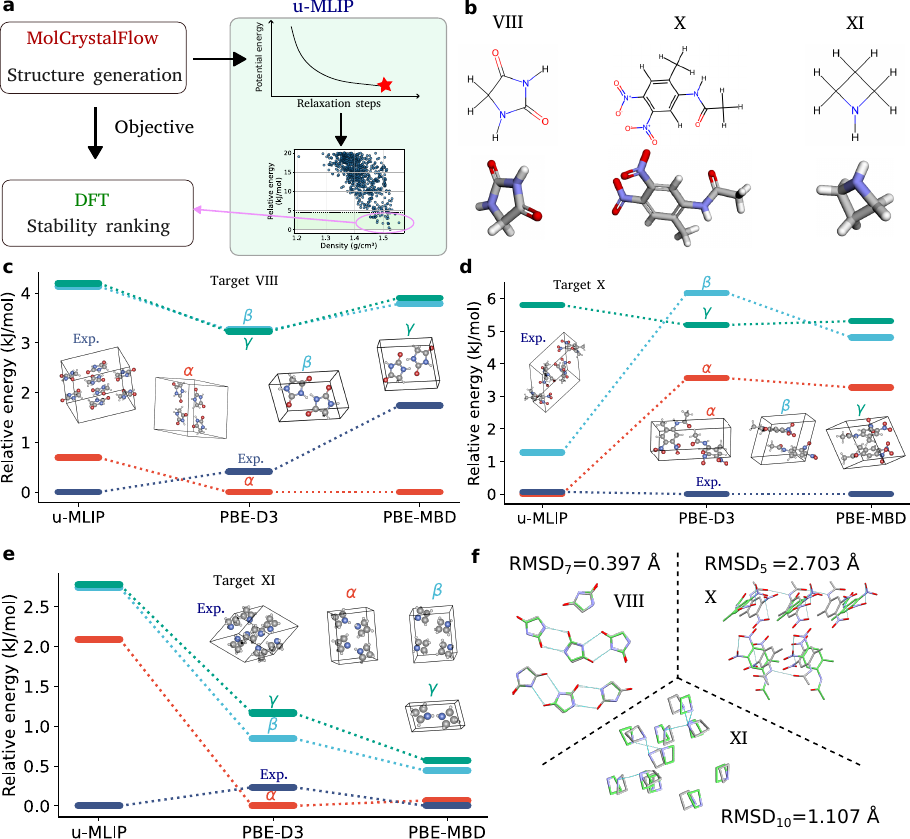}
    \caption{\textbf{Molecular crystal structure prediction by integrating MolCrystalFlow generator, u-MLIP and DFT stability ranking.} a, Four-step pipeline to integrate MolCrystalFlow for structure generation, u-MLIP for structural optimization and obtaining coarse energy landscape, and DFT for final stability ranking. b, Molecular graphs and conformer structures of three open targets from the 3$^{\mathrm{rd}}$ CCDC CSP competition. c--e, Energy ranking results evaluated with u-MLIP, PBE-D3 and PBE-MBD for three generated and relaxed polymorphs whose PBE-MBD energies are the lowest, combined with the experimental structure evaluated at the same level of theory. Figure c, d and e correspond to target VIII, X and XI, respectively. f, Packing similarity analysis of the PBE-MBD lowest-energy polymorph relative to the experimental structure was performed using the COMPACK algorithm as implemented in the CSD Python API. 
    }
    \label{fig:csp}
\end{figure}

To assess MolCrystalFlow under a realistic CSP scenario, we construct a pipeline that integrates the MolCrystalFlow generator, a universal machine-learning interatomic potential (u-MLIP), and density functional theory (DFT) calculations (Figure~\ref{fig:csp}a).
We evaluate this workflow on three publicly released targets from the $3^{\mathrm{rd}}$ CCDC CSP Blind Test~\cite{day2005} (Figure~\ref{fig:csp}b). 
All three experimental crystal packings contain two different axis-flip states with equal numbers of molecular building blocks. 
The iodine-containing target IX is excluded, as iodine is not present in the training data.
The CSP workflow consists of two stages: structure generation and structure ranking. 
Structure generation itself consists of two steps, namely conformer generation followed by crystal packing generation. 
MolCrystalFlow targets the crystal packing generation stage by efficiently sampling putative polymorphs. 
However, these generated structures are not fully optimized. 
Structural optimization is therefore critical for reliable ranking, as competing polymorphs, once relaxed, often differ in lattice energies by only a few kJ/mol, on the order of chemical accuracy (4.18 kJ/mol).
To avoid prohibitively expensive large-scale DFT evaluations, we employ the Universal Model for Atoms (UMA), specifically the 6M uma-s-1.1 model with the OMC task, as an intermediate screening tool~\cite{wood2025}. 
An initial energy landscape is constructed using u-MLIP through a two-stage relaxation procedure (Figure~\ref{fig:energy_landscape}). 
From this landscape, we select the top-10 structures with relative lattice energies closest to the minimum for subsequent DFT evaluation.
This two-stage relaxation procedure is detailed in \hyperlink{sec:computational_settings}{`Computational settings'} section.

Using the top-10 candidates selected by the UMA-OMC model, we performed more rigorous ranking using 0K DFT energies at the levels of PBE-D3 and PBE-MBD.
While these dispersion-corrected functionals are widely used, it is important to note that quantitatively accurate lattice energy prediction for molecular crystals remains challenging even for state-of-the-art DFT methods~\cite{perry_taming_2025, chattopadhyay_lattice_2025}.
For each target, we identify three representative polymorphs, denoted by $\alpha$, $\beta$ and $\gamma$ ordered by their PBE-MBD energies.
Their relative energy rankings, together with the experimental structure, are shown at the levels of UMA-OMC u-MLIP, PBE-D3 and PBE-MBD (Figure~\ref{fig:csp}c--e for targets VIII, X and XI, respectively).
Overall, despite discrepancies in the exact ordering, u-MLIP captures the general ranking trend when compared to DFT methods.
For target VIII and XI, both PBE-D3 and PBE-MBD identify a four-molecule ($Z=4$) crystal to be close in energies to the experimental structures ($Z=8$).
For target X, UMA-OMC suggests a $Z=2$ polymorph close to the experimental $Z=4$ structure, but DFT places it approximately 3 kJ/mol higher in relative energy.
These 0 K energy rankings indicate that MolCrystalFlow effectively generates molecular crystal structures that reside within low-energy basins of energy landscapes.

Beyond energy ranking, we evaluate packing similarity for the PBE-MBD lowest-energy polymorphs versus experimental structures using the COMPACK program implemented in CSD Python API~\cite{groom2016}. 
This program aims to maximize the number of matched molecules and reports the root-mean-square deviation (RMSD) over the matched cluster.
Consistent with the energy results, target VIII exhibits seven matched molecules with $\mathrm{RMSD}_7 = 0.397$~\AA, and target XI shows $\mathrm{RMSD}_{10} = 1.107$~\AA.
For both targets, the predicted structures produce similar hydrogen-bonding networks compared to experiment, although the lowest-energy polymorph for target XI show more distinct packing differences.
For target X, the model fails to recover a comparable hydrogen-bonding network ($\mathrm{RMSD}_5 = 2.703$~\AA), highlighting the difficulty of identifying the exact experimental polymorph within a complex energy landscape.
Details of DFT calculations and the COMPACK program are provided in the \hyperlink{sec:computational_settings}{Computational settings} section.

We further applied Genarris-3 to the same three targets. 
Its energy landscapes (Figure~\ref{fig:genarris_energy_landscape}) show a smaller fraction of structures within 20 kJ/mol of the minimum compared to MolCrystalFlow, although for target XI it identifies more structures within 2.5 kJ/mol of the minimum. 
With space-group constraints, Genarris-3 successfully recovers the experimental structure for target VIII, but struggles for targets X and XI (Figure~\ref{fig:genarris_rmse}), resulting in performance comparable to MolCrystalFlow.
Overall, this CSP study demonstrates the potential of integrating a flow-based generative model with universal machine-learning interatomic potentials to accelerate the discovery of stable molecular crystal polymorphs. 
At the same time, the difficulty in consistently reproducing experimental crystal structures underscores the remaining challenges and opportunities for further methodological advances.

\section*{Discussion}
MolCrystalFlow leverages a hierarchical representation together with Riemannian flow matching to accelerate molecular crystal structure prediction. 
By efficiently sampling densely packed molecular crystals, it achieves state-of-the-art performance among flow-based models for large periodic systems and delivers competitive performance relative to multi-shot, rule-based crystal structure prediction methods.

Despite these advances, several limitations remain. First, the current model is fundamentally data-driven and trained solely on structural information, without explicit access to energetic information. 
As a result, generating intrinsically low-energy crystal packings still remains challenging. 
This limitation is common to many data-driven generative models and may be alleviated by incorporating energy-based formulations~\cite{liu2025} or coupling generative sampling with inference-time scaling and reweighting strategies~\cite{mark_feynman-kac-flow_2025}.

In addition, although open molecular crystal datasets used in this study satisfy rigid-body assumptions, conformational polymorphism plays a decisive role in many experimentally observed crystals~\cite{groom2016, bauer2001}. 
Extending MolCrystalFlow to include torsional degrees of freedom would allow the model to capture intramolecular flexibility without resorting to fully all-atom representations~\cite{kim2025-2}. 
Such an extension would strike a balance between physical fidelity and computational tractability.
Finally, experimental molecular crystals often crystallize in a limited subset of space groups, and explicitly leveraging space-group symmetry constraints could substantially reduce the generative search space~\cite{groom2016}.
Operating directly on space-group–constrained manifolds~\cite{puny2025} or reduced asymmetric unit representations~\cite{levy2024} represents a promising direction for improving both efficiency and accuracy.

Although molecular crystal structure prediction remains an open and challenging problem, MolCrystalFlow provides an important step toward end-to-end generative design of molecular crystals. By integrating hierarchical representations and Riemannian flow matching, it lays the groundwork for future advances in molecular crystal structure prediction based on generative models.

\section*{Methods\label{sec:methodology}}
\newcommand{\p}{\mathbf{p}}
\newcommand{\bigsquarebracket}[1]{\left[#1\right]}
\newcommand{\bigcurlybracket}[1]{\left\{#1\right\}}
\newcommand{\bigbracket}[1]{\left(#1\right)}
\newcommand{\norm}[1]{\lVert #1 \rVert}
\newcommand{\aset}[1]{\{#1\}}
\newcommand{\nprod}{\prod_{i=1}^N}
\newcommand{\nsum}{\sum_{i=1}^N}
\renewcommand{\thefootnote}{\roman{footnote}}

\subsection*{Hierarchical Representation of Molecular Crystals}\label{sec:representation}

\paragraph{Invariant embedding of individual molecules.}
We represent a molecule as a set of atoms with coordinates $x = (x_1,\dots,x_N) \in \mathbb{R}^{N\times 3}$ and invariant atomic embedding $h = (h_1,\dots,h_N) \in \mathbb{R}^{N\times n_f}$, where $N$ is the number of atoms in the molecular building block. 
The atomic embeddings remain invariant under the E(3) coordinate transformation including translations, rotations, and reflections. 

To embed the rigid molecular building block (BB), we use an equivariant graph neural network (EGNN). Node embeddings are updated through message passing between atoms within a cutoff radius:
\begin{equation}\label{eq:egnn_message_passing}
    m_{ij} = \phi_m(h_i^l, h_j^l, d_{ij}^2, \phi_d(d_{ij})), \quad h_i^{l+1} = \phi_h(h_i^l, \sum_{j \neq i} m_{ij}),
\end{equation}
where $l$ indexes the layers, $d_{ij} = \|x_i - x_j\|$ is the euclidean distance between atom $i$ and $j$, and $\phi_d(d_{ij})$ is a Gaussian-expanded distance encoding detailed in Section~\ref{sec:rbf_embedding}. $\phi_e$ and $\phi_h$ denote edge and node multilayer perceptrons (MLPs), respectively. Because the message function depends only on distances, the embedding is E(3)-invariant.
After message passing, the BB embedding is computed as a learnable weighted mean of the final node embeddings:
\begin{equation}\label{eq:weighted_mean_embedding}
\hat{h}_{BB} = \sum_i w_i \cdot h_{i,\mathrm{final}}, \qquad w_i=\phi_w([h_{i, final} \oplus  \| x_i \|]), 
\end{equation}
where the weights $w_i$ are predicted from a shallow MLP taking input as the concatenated final node features and each atom’s distance to the molecular centroid.

To enrich the representation, we additionally incorporate auxiliary molecular-level descriptors $\psi$, inspired by Kilgour~\cite{kilgour2023}. 
These descriptors include basic, chemical and geometric features.
Three basic features consist of total atom count, number of heavy atoms and molecular weight.
Eight chemical features are obtained through RDkit~\cite{rdkit}, comprising an indicator of chirality, counts of hydrogen bond donors acceptors, number of rotatable bonds, number of aromatic rings, octanol-water partition coefficient ($\log{P}$), and topological polar surface area  (TPSA). 
Seven geometric features include the radius of gyration, asphericity, eccentricity,  planarity and three lengths of each principal components. 
The combined 18 auxiliary descriptors are concatenated with the invariant EGNN-derived BB embedding to form the final building-block representation:
\begin{equation}\label{eq:final_bb_embedding}
h_{\rm{BB}}=[\hat{h}_{\rm{BB}} \oplus \psi],
\end{equation}
which is the input representation in the next flow stage that learns molecular packings in the crystal.

\paragraph{Molecules in the crystal.}
For flow matching, each molecule is represented as a rigid body characterized by a centroid fractional coordinate ($F_{\mathrm{BB}} \in \mathbb{R}^3$) and a rotational orientation ($R_{\mathrm{BB}} \in \mathbb{R}^{3 \times 3}$). 
We represent molecule centroids in fractional coordinates to enforce lattice periodicity and avoid discontinuities when centroid positions cross lattice boundaries during generation.
$F_{\mathrm{BB}}$ is defined as the mean of the atomic coordinates transformed by by an inverse of lattice matrix, and the orientation is obtained by performing principal component analysis (PCA) on the atomic positions subtracted by a centroid position.
To ensure a consistent choice of axes and preserve a right-handed coordinate system, we first align the PCA eigenvectors using an equivariant reference vector, followed by a normalization process to ensure that each molecular building block shares the same intramolecular atomic coordinates, which removes the ambiguity in SO(3) geodesic paths for isomorphic right-handed principal axes.   

Combined with a lattice matrix $L \in \mathbb{R}^{3 \times 3}$, where $\det(L) > 0$, $F_{\mathrm{BB}}$ and $R_{\mathrm{BB}}$  specify the global arrangement of molecules in the crystal. 
The intra-molecular atomic coordinates $x_{i,\mathrm{PCA}}$ are defined in the molecule’s PCA frame and remain fixed throughout the flow matching stage.
They are used only when reconstructing the final all-atom crystal structure from $(F_{\mathrm{BB}}, R_{\mathrm{BB}}, L)$.
Under the rigid-body approximation, a molecular crystal is therefore specified by
\begin{equation}\label{eq:mc_representaion}
    \{(F_{i, \rm{BB}}, R_{i, \rm{BB}})\}^n_{i=1},\quad L,
\end{equation}
where $n$ is the number of molecules (building blocks) in the crystal. 
For each building block $\mathcal{M}_{\mathrm{BB}}$, we additionally retain its internal coordinates and atom types $\mathcal{M}_{\mathrm{BB}} = (X, A)$, where $X = \{X_{j,\mathrm{PCA}}\}_{j=1}^m$ are PCA-aligned atomic positions, $A = \{A_j\}_{j=1}^m$ are the corresponding element types, and $m$ is the number of atoms in the molecule.

 Thus, the crystal is parameterized by
\begin{equation}\label{eq:fractional_coordinates}
\mathcal{C} = (F, R, L),
\end{equation}
where $F \in [0,1)^{3 \times n}$ contains the fractional coordinates of all building block centroids, $R \in \mathbb{R}^{3 \times 3 \times n}$ stores their rotation matrices, and $L$ is the lattice matrix.
This formulation separates local molecular geometry from global crystal packing and is therefore well-suited for scalable generative modeling.

\subsection*{Extracting the axis-flip state $\chi$}\label{sec:extract_chi}

For each molecular building block, we define an equivariant local reference from its atomic coordinates and species based on the previous works of Gao~\cite{gao2023samplingfree} and Kim et al.~\cite{kim2025}. 

\paragraph{PCA axes.} Consider a molecular building block with atomic coordinates $\{X_i \in \mathbb{R}^3\}_{i=1}^N$ where $N$ is the number of atoms in the molecule and atomic masses as $\{m_i\}_{i=1}^N$. 
We first compute the geometric centroid $\bar{X} = \frac{1}{N} \sum_{i=1}^N X_i$.
The centered coordinates are denoted as $Y_i = X_i - \bar{X}$.
Principal axes for the rotational orientation of building blocks are obtained via an eigendecomposiiton of the covariance matrix $\mathrm{Cov}(Y) = \frac{1}{N} \nsum Y_i Y_i^T$. 
The eigenvectors are then identified and ordered by the eigenvalues to form a rotation matrix $U = [u_1, u_2, u_3]$, with the corresponding eigenvalues in the order of $\lambda_1 \ge \lambda_2\ge \lambda_3$.

\paragraph{Equivariant displacement vector.}
Consider the center of mass for the molecule $\bar{X}_\mathrm{CM}= \sum_i m_i X_i /\sum_i m_i$. 
An equivariant displacement vector is then constructed as $D = \bar{X}_\mathrm{CM} - \bar{X}$.
If $D=0$ for symmetric molecules, it is replaced by the displacement vector connecting the geometric centroid to its nearest neighbor.

\paragraph{Binary axis-flip state $\chi$.} We project $D$ onto the PCA axes, which gives us $\nu_k = u_k^T D$, $k=1,2,3$. To remove the ambiguities of PCA  we define the discrete axis-flip vector $\eta = \left(\mbox{sign}(\nu_1), \mbox{sign}(\nu_2), \mbox{sign}(\nu_3) \right) \in \{-1, +1\}^3$. 
To remove symmetry-induced degeneracies, we introduces a indicator for the number of axes flipped with respect to $D$: $n_{-} = \sum_{i=1}^3 \mathbb{I}[\eta_k = -1]$. The unique axis-flip state $\chi$ is hence given by
\begin{equation}\label{eq:chi}
\chi = \left\{ \begin{array}{rcl} 
0 & \mbox{for} & n_{-}=0,2 \\
1 & \mbox{for} & n_{-}=1,3 
\end{array}\right.,
\end{equation}
To ensure consistency of local coordinates across symmetry-equivalent building blocks, we renormalize local coordinates and rotation matrices based on the axis sign states so that local coordinates within the same $\chi$ group axis-flip states are the same and differ by an opposite sign for one axis across different $\chi$ groups. 
This procedure removes spurious axis inversions while preserving SE(3) equivariance and consistent local frame coordinates across different building blocks.

\subsection*{Geometric Symmetry of Molecular Crystal Structures}\label{sec:geometric_symmetry}
The molecular crystal distribution $p(L, F, R \mid h_{\rm{BB}}(X, A))$ should satisfy geometric symmetries. 
For brevity, we drop the conditioning on the building-block embeddings and write the target distribution as $p(L, F, R)$.
Geometric symmetries of inorganic crystals and molecular systems have been widely studied in prior work~\cite{kohler2020, jiao2024, xu2022}. 
In our setting, the distribution respects permutation invariance, periodic translational invariance, and $SO(3)$ invariance.
Permutation invariance requires that reordering the same types of building blocks does not change the distribution, which is naturally handled when the generative model is parameterized using a graph neural network. $SO(3)$ invariance ensures that a global rotation of the entire crystal leaves the distribution unchanged.
In other words, for any orthogonal transformation $Q \in \mathbb{R}^{3 \times 3}$ where $Q^T Q=I$, we require $p( LQ^T, F, Q R) = p(L, F, R)$. 
A right matrix multiplication is used for the lattice matrix since we adopt a row-vector convention for the lattice matrix throughout this work.
Because fractional coordinates are defined relative to the lattice basis, they remain unaffected by such operations. 
Periodic translational invariance reflects the fundamental periodicity of crystals: translating all building-block centroids by any lattice vector should correspond to the same configuration. This is expressed as $p(L, \tau(F + T \cdot \mathbf{1}^\top), R) = p(L, F, R)$, where $T$ is a translation vector, $\mathbf{1}$ is a vector of ones, and $\tau(\cdot)$ wraps coordinates back into the unit cell.
These symmetries guide the design of the generative model so that the learned distribution is physically meaningful and consistent with the underlying geometry of molecular crystals.
One should however note there is an ongoing debate whether the geometric symmetries can be comprised by simple extensive data augmentation on non-equivariant neural network architectures~\cite{jin2025oxtal, bigi2026, batzner2022}.  

\subsection*{Joint Flow Matching on Riemannian manifolds}\label{sec:joint_flow_matching}

We employ a Riemannian flow matching generative process to learn the joint distribution  $p(L,F,R)$.
A conditional velocity field and an interpolant is defined between pairs of samples drawn from the base and data distribution. 
This conditional velocity field is constructed directly on the Riemannian manifolds for key molecular crystal modalities, which ensures that numerical integration of the learned velocity fields produce valid states that remain on the manifold. 
The flow matching process is parameterized by a period E(3)-invariant graph neural network detailed in the subsequent section.

The molecular centroids are represented by fractional coordinates which are wrapped around the range of [0, 1), hence living on three-dimensional tori for 3D centroid positions.
The orientation can be intuitively represented by a 9-element rotation matrix on an $SO(3)$ manifold.
Yet, entries of the 9-element matrix are not independent; to directly learn the flows of orientations, the rotation matrix is converted to a 3-element axis-angle representation, which is a reparameterization of the $SO(3)$ manifold through the Rodrigues' rotation formula, and any arbitrary vector in this representation corresponds to a valid rotational orientation.
Despite enforcing the constraint $\det(L) > 0$, the lattice matrix is further normalized to a lower-triangular form with lattice lengths ordered as $a \le b \le c$. This standardization removes redundant lattice representations and simplifies the learning of lattice matrices.
Lattice matrices are represented in its Euclidean space with informed base distributions fit from the training data lattice parameter distribution, as used in FlowMM~\cite{miller2024} and OMatG~\cite{hoellmer2025}.

MolCrystalFlow then performs flow matching jointly over three structural modalities, including lattice matrices, fractional coordinates, and molecular orientations on their corresponding manifolds~\cite{chen2024}. 
A simple linear interpolant connecting each modality between $t=0$ and $t=1$ is used on its geodesic path, giving rise to the following forms:
\begin{equation}\label{eq:rm_geodesics}
L_t =(1-t) L_0 + t L_1, \qquad F_t = \tau \left((1-t) F_0 + t F'_1 \right), \qquad R_t = \exp_{R_0}\left(t \log_{R_0} (R_1)\right),
\end{equation}
where $F'_1$ indicates a periodic image by unwrapping the end state $F_1$ in order to find the shortest geodesic path between $F_1$ and $F_0$, and $\tau(\cdot)$ wraps fractional coordinates back to the torus. Differentiating the interpolants yields the conditional velocity fields, which define the training objective: 
\begin{equation}\label{eq:rm_geodesics_velocity}
 u_t(L_t | L_1, L_0) = \frac{L_1-L_t}{1-t}, \qquad  u_t(F_t|F_1, F_0) = \frac{F'_1 - F_t}{1-t}, \qquad u_t(R_t|R_1, R_0) = \frac{\log_{R_t}(R_1)}{1-t}.
\end{equation}
The loss combines denoising end-point predictions for lattices and rotations, along with a velocity-field objective for fractional coordinates,
\[
\mathcal{L}(\theta)=\mathbb{E}_{t,(\mathcal{C}_0, \mathcal{C}_1})\left[\frac{1}{(1-t)^2}\left(\lambda_L\|\hat{L}_1-L_1\|^2 
+ \lambda_R \|\log_{R_t}(\hat{R}_1)-\log_{R_t}(R_1)\|^2\right)
+ \lambda_F \|\hat{u}_t(F_t)-u_t(F_t)\|^2\right].
\]
We set $\lambda_L = 0.1$, $\lambda_R = 1$, and $\lambda_F = 2$ in this work.
To prevent numerical instability near $t=1$ during training, we clip time at $t=0.9$; that is, we transform the denominator $(1-t)$ in the loss function to $\left(1-\min(t, 0.9)\right)$ to avoid a division by zero. 

\subsection*{Neural Network Architecture for Periodic E(3)-Invariant Flow Matching}\label{sec:nn_arc}
We parameterize the joint flow matching over lattice matrices, rotation matrices, and centroid fractional coordinates using a periodic E(3)-invariant graph neural network adapted from the CSPNet~\cite{jiao2024}. 
The model learns the velocity fields of fractional coordinates ($u_t(F_t)$), rotational orientation ($u_t(R_t)$) and lattice matrices ($u_t(L_t)$).
The initial node embeddings are given by the building block (BB) representations, concatenated with a time embedding for $t$ and the projected into the initial dimension of the BB embedding. 
During message passing, interactions between two building blocks depend on their relative fractional coordinates, relative orientations, and the current lattice matrix.
The lattice matrix $L_t$ is shared across the entire crystal, while the fractional coordinates $F_t$ and the rotation matrices $R_t$ are defined for each building block. At layer $s$, message passing between building blocks $i$ and $j$ is defined as:

\begin{align}\label{eq:fm_message_passing}
m_{ij}^{(s)} &= \phi_m \left(h_i^{(s-1)}, h_j^{(s-1)}, L^\top L, \psi_{\rm FT}(F_j - F_i), \Psi_R(R_i^\top R_j), m_{R, L}, m_{F, L}\right), \\
m_i^{(s)} &= \sum_{j=1}^N m_{ij}^{(s)}, \quad
h_i^{(s)} = h_i^{(s-1)} + \phi_h\left(h_i^{(s-1)}, m_i^{(s)}\right),
\end{align}
where $\psi_{\mathrm{FT}}$ is a Fourier-based embedding using trigonometric functions enforcing periodicity on the torus~\cite{vaswani_attention_2017}, and $\Psi_R(R_i^\top R_j)$ encodes relative orientations.
The matrix $R_i^\top R_j$ lies in $SO(3)$ because we have $\det(R_i^\top R_j)=1$.
Therefore, it can be expressed in an axis–angle representation. We compute the rotation angle $\omega$ and the azimuthal and inclination angles $\rho$ and $\kappa$. Since $\omega$ and $\theta$ are periodic (mod $2\pi$), we embed $\omega/2\pi$ and $\theta/2\pi$ using trigonometric functions, while $\kappa$ is encoded via a shallow MLP $\phi(\cdot)$:
\begin{equation}\label{eq:so3_ft}
\Psi_R(R_i^T R_j) = [\psi_{\rm{FT}}(\omega/2\pi) \oplus \psi_{\rm{FT}}(\rho/2\pi) \oplus \phi(\kappa)].
\end{equation}
Regarding the interaction terms of $L$ with $R$ and $F$, we have unit vectors $m_{R,L}=L^TLR_i^T R_j/\lVert L^TLR_i^T R_j + \epsilon\rVert$ and $m_{F,L}=L^TL(F_j - F_i)/\lVert L^TL(F_j-F_i) + \epsilon\rVert$ where $\epsilon=10^{-8}$ is to avoid divide-by-zero.
All inputs to message passing are thus invariant to global rotations and translations, making the network periodic E(3)-invariant.
After $f$ message-passing layers, the final node embeddings $h_i^{(f)}$ are used to parameterize the outputs:
\begin{align*}
\Delta R_i &= \phi_R \left(h_i^{(f)}\right), \quad R_{i, t=1} = R_i \times \Delta R_i \\
\Delta L_j &= \phi_{L_j} \left( \frac{1}{m}\sum_{k \in \mathcal{C}_j} h_k^{(f)} \right) \times L_j^T, \quad L_{j, t=1} = L_j + \Delta L_j \\
u_t(F_t) &= \phi_F(h_i^{(f)})
\label{eq:final_pred},
\end{align*}
The update of the lattice is invariant to global lattice rotation, and multiplication by $L_t$ ensures equivariance, demonstrated by Jiao et al.~\cite{jiao2024}.
Similar compositional operation is applied to the rotational update to ensure the global rotational equivariance of building block orientations.

During inference, the centroid velocity field $u_t(F_t)$ is integrated directly and the predicted fractional coordinate is mapped back onto the torus.
In contrast, the velocity fields at time $t$ for lattice matrix $L$ and molecular orientations $R$ are computed based on the denoised states and interpolant states at time $t$.
The derived velocity fields are then integrated numerically in time to move to the next time step. 
The ordinary differential equations (ODEs) governing all velocity fields are solved using Euler integration with 50 discrete time steps.
Velocity annealing is applied during integration for both the fractional coordinates and rotational orientations, with scaling factors $s_{u_F}$ and  $s_{u_R}$, respectively.
The state ($x_t$) update at each step is given by $x_{t+\Delta t} = x_t + (1+s t) \cdot u(x_t) \cdot \Delta t$, where the annealing factors exert the strongest influence as $t \to 1$.

\subsection*{Base Distributions and Optimal Transport}\label{sec:base_p}

\paragraph{Base distributions of fractional coordinates, lattice matrices and rotation matrices.}  Fractional coordinates are drawn from a uniform prior with support on [0,1).
Lattice matrices are parameterized by six lattice parameters fitted from the training data and constructed in a normalized lower-triangular form.
Lattice lengths are sampled from Gaussian distributions, while lattice angles are sampled from a uniform distribution with support [60$^\circ$, 120$^\circ$].
The sampled lattice lengths and angles are then used to construct the lattice matrix base distributions in the lower-triangular form. 
For molecular orientations, we adopt a symmetric-rotation prior.
Specifically, the first rotation matrix is sampled uniformly from the SO(3) manifold, and the remaining rotation matrices are deterministically generated according to symmetry relations with respect to this reference orientation.
As shown in the Supplementary Information, this symmetric-rotation prior preserves full SE(3) invariance, which is essential for ensuring that the generated molecular crystal structures, $p(L, F, R)$, are also SE(3)-invariant when propagated through an equivariant flow-matching process.

\paragraph{Optimal transport for fractional coordinates and rotation matrices.}
To mitigate trajectory crossings during flow matching, we employ a data-dependent optimal transport procedure that aligns source and target samples, thereby improving both inference quality and sampling efficiency.
We permute building block indices within the same $\chi$ group to minimize geodesic distances in the spaces of fractional coordinates (wrapped distances) and molecular orientations (rotational angles). 
For example, given a batch of rotation matrix pairs $(R_0, R_1)$, we seek the optimal permutation within each $\chi$ group that minimizes the geodesic distance on $SO(3)$, which can be formulated as $\operatorname*{arg\,min}_{P_\chi} d_{RM}\left(P_\chi(R_0), R_1\right)$, where $d_{RM}$ denotes the geodesic distance between rotation matrices and $P_\chi$ represents a permutation operator acting only on indices within the same $\chi$ group. 

\subsection*{Open-Source Molecular Crystal Dataset}\label{sec:datasets}

\paragraph{Data preprocessing.}
Each molecular crystal is decomposed to constituent molecular building blocks based on atomic connectivity. 
Each building block is identified by using a graph-based traversal constructed from interatomic neighbors using element-dependent cutoff radii. 
The neighboring environment is determined through atomic simulation environment~\cite{hjorth2017}, and the cutoffs for each element is manually chosen based on covalent radius of each atom type. Two atoms are considered bonded if their separation distance falls within the add-up of a pair of atoms.
A minimum-image convention is considered to account for cross-cell distances across periodic boundaries. 
Each grouped molecule will share a building block index that is used for determining the centroids and rotational orientation.
In addition, each molecular building block is unwrapped so that all atoms are fully connected in their Cartesian coordinates.
To facilitate learning of lattice matrices, we first standardize each lattice by sorting the lattice vectors by their lengths, then transform the reordered matrix into a lower-triangular form via a `QR' decomposition applied to its transpose.

\paragraph{Thurlemann molecular crystal dataset.}
We evaluate MolCrystalFlow on the molecular crystal dataset curated by Thurlemann et al.~\cite{thurlemann2023}, which is derived from the Cambridge Structural Database~\cite{groom2016}.
The dataset comprises molecular crystals containing seven elements (H, C, N, O, F, Cl, S) and originally includes 11,489$\times$5 structures, where the five entries correspond to distinct lattice strain levels.
All structures were relaxed with a fixed experimental cell at the DFT level using the PBE-XDM functional as implemented in Quantum ESPRESSO~\cite{giannozzi2017}.
We selected the 11,489 unstrained structures as the initial candidate dataset.
After preprocessing and filtering, the final dataset consists of 11,488 structures.
The structure filtered out is one structure classified to have four building blocks but actually labeled with only two molecules in the original data source ($Z=2$). 
This structure is shown in Figure~\ref{fig:thurlemann_outlier}. 
The dataset was split such that identical chemical formulas do not appear across different splits, with 10,000, 738, and 750 structures used for training, validation, and testing, respectively.
A detailed explanatory data analysis is provided in Figure~\ref{fig:eda_thurlemann}.

\paragraph{Selected OMC25 data (OMC25-MCF).} 
OMC25 is the largest publicly available molecular crystal dataset generated at the PBE-D3 level by FAIR Chemistry using VASP~\cite{gharakhanyan2025_OMC25}.
It comprises more than 26 million structures, sampled from geometry-optimized conformers in the OE62 dataset~\cite{stuke_atomic_2020}.
The dataset contains 12 elements (C, H, N, O, P, S, Cl, F, I, B, Br, Si), and it is restricted to single-component molecular crystals and adopts a rigid-body approximation.
For training MolCrystalFlow, we retain a single crystal packing per conformer, selected as the polymorph of the lowest lattice energy with a maximum atomic force below 0.05~eV/\AA.
After filtering, a total of 46,801 structures remain, and this selected subset of OMC25 is termed `OMC25-MCF'.
We split the dataset into 42,120 training structures (90\%), 3,681 validation structures, and 1,000 test structures.
Exploratory data analysis of the full OMC25-MCF data is also provided in Figure~\ref{fig:eda_omc25}.

\subsection*{Evaluation Metrics}\label{sec:metrics}

\paragraph{Evaluation on the raw generated structures.} Crystal structure generation performance is first evaluated using \textit{StructureMatcher} from the pymatgen library~\cite{ong_python_2013}, with lattice tolerance \textit{ltol=0.3}, angle tolerance \textit{atol=10}, and site tolerances in the range of 0.5 to 1.2.
Volume scaling is disabled because uniform rescaling would alter intramolecular bond lengths, which is undesirable for molecular crystal structures. 
Owing to the presence of numerous polymorphs for a given molecular conformer, multiple candidate structures are generated for each test case. 
A prediction is considered correct if at least one generated structure matches the corresponding ground-truth structure under the above criteria.

\paragraph{Evaluation using CCDC CSP compounds.} We further benchmark MolCrystalFlow on three targets from the $3^{\mathrm{rd}}$ CCDC CSP Blind Test~\cite{day2005}. 
Among these targets, the experimental crystal packings of targets VIII and XI consist of eight molecules per unit cell ($Z=8$), whereas target X has $Z=4$.
For sampling with MolCrystalFlow, we consider unit cells containing $Z \in \{2,4\}$ molecules, which are among the most frequently observed values in the Cambridge Structural Database (CSD)~\cite{groom2016}.
Although $Z=1$ and $Z=8$ also account for a substantial fraction of molecular crystals in CSD, we do not sample these cases here. 
Preliminary tests indicate that $Z=1$ structures are less stable than those with $Z=2$ and $Z=4$, while the Thurlemann dataset contains only a limited number of $Z=8$ examples, making reliable generation of eight-molecule unit cells challenging.
For each value of $Z$, we generate 200 candidate structures under a maximum intermolecular ellipsoidal volume overlap of 5\%. 
We consider two $\chi$-state scenarios: one in which all molecules share the same $\chi$ state, and another in which the crystal contains equal numbers of molecules in two distinct $\chi$ states.
This results in a total of 800 sampled structures per molecular conformer.
The entire sampling procedure is repeated twice using different random seeds.
Structural optimization and energy ranking are performed using the UMA-OMC (6M UMA s-1.p.1 model) universal machine--learning interatomic potentials (u-MLIP), followed by DFT calculations at the PBE-D3(BJ) and PBE-MBD levels of theory.
We further examine structural fidelity by computing the root mean squared distances (RMSDs) for 15-molecule assemblies, and RMSDs are evaluated on matched molecules using the \textit{COMPACK} program as implemented in the CSD Python API~\cite{groom2016}.
Molecular matching is performed using distance and angular tolerances of 50\%, excluding hydrogen atoms and without allowing molecular differences.

\subsection*{Computational settings}\label{sec:computational_settings}

\paragraph{Model Training and Inference.}
Both MolCrystalFlow and MOFFlow models were built using PyTorch. 
Both training and inference use one Nvidia B200 graphic card with 192 GB memory.
The baseline MOFFlow model was trained using the same hyperparameter settings of the original work~\cite{kim2025}. 
Hyperparameters of MolCrystalFlow are included in Table~\ref{tab:hyperparams}.

\paragraph{u-MLIP model evaluation and structural optimization.}
UMA-OMC u-MLIP was used to optimize the directly generated structures, and it is a two-stage process implemented with atomic simulation environment (ASE)~\cite{hjorth2017}.
First, each generated structure is relaxed for up to 100 steps or until the maximum atomic force is below 0.05 eV/\AA~using a rigid-body constraint and a fixed lattice based on the BFGS algorithm. 
Second, the full atomic and lattice degrees of freedom are relaxed using the UMA-OMC model for up to 1000 steps or until the maximum atomic force falls below 0.01 eV/\AA, implemented via the \textit{FrechetCellFilter} of ASE.
The 10 lowest-energy structures across all sampled $Z$ values are selected for final DFT validation for each round.
We have in total 20 structures for two rounds with different structure generation random seeds.

\paragraph{DFT calculations.}
We used VASP 6.4.3 for DFT evaluations~\cite{kresse_efficient_1996}. 
The exchange-correlation interactions were described using Perdew–Burke–Ernzerhof (PBE) generalized gradient approximation~\cite{perdew_generalized_1996}.
Core-valence interactions were treated using the projected augmented-wave (PAW) method, with a plane-wave energy cutoff of 520 eV.
The structural relaxation was performed with the conjugate-gradient algorithm with a force convergence criterion of 0.01 eV/\AA~ and an energy convergence threshold of 10$^{-7}$ eV. 
A maximum of 1500 ionic steps was used.
Brillium-zone integrations employ the F-centered Monkhorst-Pack scheme, with a Gaussian smearing of 0.05 eV.
K-point meshes were generated automatically using the pymatgen based on a k-point density criterion of 1000 k-points per reciprocal atom~\cite{ong_python_2013}. 
Two types of dispersions were considered, including D3 (BJ)~\cite{grimme_consistent_2010} and many-body dispersion (MBD)~\cite{tkatchenko_accurate_2012}.

\paragraph{Structure generation with Genarris-3.}
We applied Genarris-3.0 \cite{yang_genarris_2025} as a baseline comparison using a two-step workflow that first generates trial molecular crystal structures across compatible space groups and then compacts each candidate with Symmetric Rigid Press while preserving space-group symmetry. For the X, XI, and VIII targets, we generated 800 accepted structures per space group, while for the Thurlemann validation set we generated 200 accepted structures per space group. In the generation step we allowed up to 10,000,000 attempts per space group and per sampled volume. Unit-cell volumes were set from the built-in volume prediction and initialized at 1.5$\times$ the predicted mean to facilitate placement before compaction. Short-range contact screening used sr = 0.85 with tol = 0.01, and neighbor identification used natural\_cutoff\_mult = 1.2. In the Symmetric Rigid Press step, structures were compressed toward close packing using a smooth overlap-penalty model optimized with BFGS to tol = 0.01, with sr = 0.85 and natural\_cutoff\_mult = 1.2 applied consistently. For the validation set, the resulting structures were uniformly subsampled to produce Figure~\ref{fig:benchmark}. 

\paragraph{Rigid press optimization for MOFFlow- and MolCrystalFlow-generated structures.}
To optimize generated structures, we first uniformly scale the lattice cell by a factor of 1.5. 
Rigid-body press optimization is then performed using the same settings described above with the space group fixed to 1. 
After optimization, the lattice volume is rescaled to match the original volume of the raw generated structures. 
Structures that fail to converge during optimization are kept in their original generated form. 
The overall success rate of rigid-body optimization is approximately 86\% for both MolCrystalFlow and MOFFlow.

\section*{Data and code availability}
Data and model checkpoints can be found at Zenodo~\cite{zenodo_molcrystalflow}. Open-source code is provided at Github: https://github.com/Liu-Group-UF/MolCrystalFlow.

\section*{Acknowledgment}
Cheng Zeng thanks Michael Kilgour (NYU) for insightful discussion during the AI for Chemistry symposium held at New York University. The authors gratefully acknowledge Amit Gupta (UMN) for assistance with the dataset during the initial stage of this project. The authors acknowledge funding from NSF Grant OAC-2311632 and from the AI and Complex Computational Research Award of the University of Florida. S.M also acknowledges support from the Simons Center for Computational Physical Chemistry (Simons Foundation grant 839534). The authors also gracefully acknowledge UFIT Research Computing for computational resources and consultation, as well as the NVIDIA AI Technology Center at UF.

\bibliographystyle{unsrt} 
\bibliography{molcrystalflow}

\end{document}